\documentclass[conference]{IEEEtran}
\IEEEoverridecommandlockouts
\usepackage{cite}
\usepackage{amsthm}
\usepackage{comment}
\usepackage{amsmath,amssymb,amsfonts}
\usepackage{algorithmic}
\usepackage{graphicx}
\usepackage{textcomp}
\usepackage{xcolor}
\usepackage{bm}
\usepackage{subfig}
\usepackage{algorithm}
\usepackage{algorithmic}

\newtheorem{theorem}{Theorem}

\newtheorem{proposition}{Proposition}

\newenvironment{shrinkeq}[1]
{ \bgroup
\addtolength\abovedisplayshortskip{#1}
\addtolength\abovedisplayskip{#1}
\addtolength\belowdisplayshortskip{#1}
\addtolength\belowdisplayskip{#1}}
{\egroup\ignorespacesafterend}

\def\BibTeX{{\rm B\kern-.05em{\sc i\kern-.025em b}\kern-.08em
    T\kern-.1667em\lower.7ex\hbox{E}\kern-.125emX}}
\begin{document}

\title{Online  Graph Learning In Dynamic Environments
}

\author{\IEEEauthorblockN{Xiang Zhang}
\IEEEauthorblockA{\textit{School   of   Information   Science   and   Engineering} \\
\textit{Southeast   University}\\
Nanjing, China \\
xiangzhang369@seu.edu.cn}
\and
\IEEEauthorblockN{Qiao Wang}
\IEEEauthorblockA{\textit{School   of   Information   Science   and   Engineering} \\
\textit{Southeast   University}\\
Nanjing, China \\
qiaowang@seu.edu.cn}

}

\maketitle

\begin{abstract}
Inferring the underlying graph topology that characterizes structured data is pivotal to many graph-based models when  pre-defined  graphs are not available.  This paper focuses on learning graphs in  the case of sequential data  in dynamic environments. For sequential data, an online version of classic batch graph learning method is developed. To better track graphs in dynamic environments, we assume graphs evolve in certain patterns  such that dynamic priors might be embedded in the online graph learning framework. When we have no knowledge of hidden patterns, a data-driven method is leveraged to  predict the evolution of graphs. Furthermore,  dynamic regret analysis of the proposed method   is  performed,  illustrating theoretically that proper dynamic priors do reduce regret. Experimental results  support  that our method is superior to the state-of-art method.

\end{abstract}

\begin{IEEEkeywords}
Graph learning, online learning, dynamic environments, regret analysis
\end{IEEEkeywords}

\section{Introduction}
\label{sec:introduction}
The goal of graph learning task is to infer the hidden topology  inside high dimensional  data \cite{mateos2019connecting,kalofolias2016learn, dong2016learning,dong2019learning}, from which a lot of  graph based tasks such as spectral clustering \cite{liu2018spectral} could be performed with a structured framework. Classic graph learning models contain  statistical models \cite{yuan2007model} and causal dependencies based models \cite{shen2016nonlinear}. With the rise of graph signal processing (GSP) \cite{ortega2018graph}, there occur vibrant methods of learning graphs from perspective of signal processing. Many GSP based models are based on  smoothness assumption \cite{dong2016learning}, which is also taken into account in this paper.

Most of the aforementioned models are of batch manner, i.e., all data are collected in advance and used totally at once to learn an optimal graph. However, this scheme does not meet many practical applications. Firstly,  it is not reasonable to learn a single graph for all data since graphs may not be static due to dynamic environments. Furthermore, the  data are often available in a sequential way. Thus  we need to update an initial inferred graph based on partial observations, and then progressively modify it as long as new data arrives. One example is to infer connections between users through shopping behavior, in which data is  generated only after each user purchases, and shopping websites hope to update the graphs of connections of users after each purchase happens. 

To this end, we need to learn time-varying graphs in an online fashion. Some excellent explorations have been made on these problems, e.g., causality based models \cite{zaman2020dynamic,zaman2020online, money2021online} and GSP based models \cite{saboksayr2021online,shafipour2019online}. However, to our understanding, most of these models ignore the  hidden evolutionary patterns of time-varying graphs, which might be useful for improving learning performance. Recent notable work \cite{Natali2021online} tries to learn Gaussian graphical model with an extra prediction step, which, however, is different from our smoothness based model.

This paper endeavors to develop a novel online graph learning framework that utilizes dynamic priors to better track time-varying graphs. We first cast batch model based on smoothness assumption into an online version via  online  mirror  descent (OMD) framework \cite{hazan2019introduction} to handle sequential data. In the situation of dynamic environments, we assume that graphs evolve in a prior dynamic pattern. We integrated the dynamic priors into the online graph learning framework as a prediction step to output graphs of the next time slot. When no explicit patterns are available, a data-driven method is leveraged to predict the evolution of graphs. We also analyze the dynamic regret, an important metric to evaluate online algorithms, of our methods to theoretically illustrate the power of dynamic priors. Numerical experiments on synthetic and real-world data show the superiority of our framework.

\section{Problem Formulation}
\label{sec:problem formulation}
\subsection{Batch Graph Learning}
Batch graph learning methods first collect $T$ observed signals $\mathbf{x}_1,\ldots,\mathbf{x}_T \in \mathbb{R}^d$  generated from graph $\mathcal{G}$, which is undirected and with nonnegative edges. One infers the topology of $\mathcal{G}$, e.g., adjacency matrix $\mathbf{W}\in \mathbb{R}^{d\times d}$ using all collected data globally at once. For smoothness based models, it is equivalent to solving the following problem \cite{kalofolias2016learn}
\begin{shrinkeq}{-1ex}
\begin{align}
    \underset{\mathbf{W}\in \mathcal{W}}{\mathrm{min}}\; \lVert \mathbf{W} \circ \mathbf{Z} \rVert_1   -\alpha \mathbf{1}^{\top}\mathrm{log}(\mathbf{W}\mathbf{1}) + \frac{\beta}{2}\lVert \mathbf{W}\rVert_{\mathrm{F}}^2,
    \label{GL-1}    
\end{align}
\end{shrinkeq}
where $\circ$ is Hadamard product and $\mathbf{1} = [1,\ldots,1]^{\top}\in \mathbb{R}^{d}$ is a column vector of ones. Moreover, $\alpha$ and $\beta$ are predefined constant parameters. Usually, the set $\mathcal{W}$ is defined as below,  
\begin{shrinkeq}{-1ex}
\begin{align}
        \mathcal{W} = \left\{ \mathbf{W} : \mathbf{W}\in \mathbb{R}^{d\times d}_+,  \mathbf{W} = \mathbf{W}^{\top},\mathrm{diag}(\mathbf{W}) = \mathbf{0}\right\},
    \label{GL-2}
\end{align}
\end{shrinkeq}
where $\mathbb{R}_+$ is the set of nonnegative real numbers and $\mathbf{0} \in \mathbb{R}^d$ is a vector with all entries equal to zero. Pairwise distance matrix $\mathbf{Z}\in \mathbb{R}^{d\times d}$ in \eqref{GL-1} is calculated using data matrix $\mathbf{X} \in \mathbb{R}^{d\times T} = [\mathbf{x}_1,\ldots,\mathbf{x}_T] = [\tilde{\mathbf{x}}_1,\ldots,\tilde{\mathbf{x}}_d]^{\top}$,
\begin{shrinkeq}{-1ex}
\begin{align}
        \mathbf{Z}_{[ij]} = \lVert \tilde{\mathbf{x}}_i - \tilde{\mathbf{x}}_j\rVert_2^2,
    \label{GL-3}   
\end{align}
\end{shrinkeq}
where $\mathbf{Z}_{[ij]}$ is the $(i,j)$ entry of $\mathbf{Z}$.
The first term of \eqref{GL-1} is used to measure the smoothness of  the observed signals. Besides, the  second  and third term control  degrees  of  each node  and sparsity  of  edges \cite{kalofolias2016learn}. Observing \eqref{GL-2}, we can find that the number of free variables of $\mathbf{W}$ is  $p=\frac{d(d-1)}{2}$. We thus define a vector $\mathbf{w}\in \mathbb{R}^p$ whose entries are the upper right variables of $\mathbf{W}$. Problem \eqref{GL-1} can then be rewritten as \cite{kalofolias2016learn}
\begin{shrinkeq}{-1ex}
\begin{align}
         \underset{\mathbf{w}\in\mathcal{W}_v}{\mathrm{min}} f(\mathbf{w}) =  \underset{\mathbf{w}\in\mathcal{W}_v}{\mathrm{min}}\,\, 2\mathbf{z}^{\top}\mathbf{w} -\alpha\mathbf{1}^{\top}\mathrm{log}(\mathbf{S}\mathbf{w}) + \beta\lVert \mathbf{w} \rVert_2^2,
    \label{reformulation-of-GL-W}
\end{align}
\end{shrinkeq}
where $\mathbf{S}$ is a linear operator satisfying $\mathbf{S}\mathbf{w} = \mathbf{W}\mathbf{1}$  and $\mathbf{z}$ is the vector form of the upper triangle variables of $\mathbf{Z}$. Additionally, the set $\mathcal{W}_v$ is defined as $\mathcal{W}_v \triangleq \{\mathbf{w}:\mathbf{w}\in \mathbb{R}^p, \,\mathbf{w}_{[i]}\geq 0, \,\mathrm{for}\,\,i = 1,..., p\}$.

\subsection{Online Graph Learning Using Dynamic Priors}
Different from  batch method waiting for all $T$ data to be ready, data arrive sequentially in online setup and we are required to update a new graph  after each data $\mathbf{x}_t, t = 1, ..., T$, is received. Furthermore, in dynamic environments, graphs are time-varying and we assume graphs evolve with a dynamic model $\Phi$, i.e.,
\begin{shrinkeq}{-1ex}
\begin{equation}
    \begin{aligned}
        \mathbf{w}_{t+1} = \Phi(\mathbf{w}_{t}),
    \end{aligned}
    \label{state transition function}
\end{equation}
\end{shrinkeq}
where $\mathbf{w}_{t}$ represents the graph at moment $t$. The model $\Phi$ describes  evolutionary patterns of dynamic graphs and may provide additional information to help track time-varying graphs. To conclude, the problem we focus on can be formally stated as follows. At each time $t$, given smooth signal $\mathbf{x}_t$  and historical data $\mathbf{x}_1,...,\mathbf{x}_{t-1}$, we are required to update a new graph $\mathbf{w}_{t+1}$ immediately by leveraging information from  $\Phi$.

\section{Proposed  algorithm}
\label{sec:Proposed Framework}
We leverage online mirror descent (OMD) algorithm  \cite{hazan2019introduction} to cast the classic batch graph learning  problem \eqref{reformulation-of-GL-W} into an online version. The available data at $t$ are $\mathbf{x}_1,...\mathbf{x}_t$ and we utilize information of the data in an recursively way. To be specific, we first calculate $\mathbf{Z}_t$ using $\mathbf{x}_t$ through \eqref{GL-3} and reshape $\mathbf{Z}_t$ as $\mathbf{z}_t$. We then update
\begin{shrinkeq}{-1ex}
\begin{equation}
    \bar{\mathbf{z}}_t = \gamma \bar{\mathbf{z}}_{t-1} + (1-\gamma)\mathbf{z}_t,
    \label{update_z_t}
\end{equation}
\end{shrinkeq}
where $\gamma \in [0,1)$ is the forgetting factor and  $\bar{\mathbf{z}}_0 $ is supposed to be $\mathbf{0}$. Note that $\bar{\mathbf{z}}_t$ carries information from $\mathbf{x}_1,...,\mathbf{x}_{t}$ and  is treated as the input data vector at $t$. With $\bar{\mathbf{z}}_t$, we can define 
\begin{shrinkeq}{-1ex}
\begin{equation}
     {f}_t(\mathbf{w};\mathbf{x}_1,...,\mathbf{x}_{t}) \triangleq  2\bar{\mathbf{z}}_t^{\top}\mathbf{w} -\alpha\mathbf{1}^{\top}\mathrm{log}(\mathbf{S}\mathbf{w}) + \beta\lVert \mathbf{w} \rVert_2^2.
    \label{def:GL-formulation-online3}
\end{equation}
\end{shrinkeq}
For notational simplicity, we omit ${\mathbf{x}}_1,...{\mathbf{x}}_t$ of $f_t$ in the remaining of the paper. Under OMD framework and given $\mathbf{w}_t$, we immediately update a new graph $\breve{\mathbf{w}}_{t}$ through
\begin{shrinkeq}{-1ex}
\begin{equation}
    \begin{aligned}
     \breve{\mathbf{w}}_{t} = \underset{\mathbf{w}\in\mathcal{W}_v}{\mathrm{argmin}}\,\, \langle\nabla  {f}_t(\mathbf{w}_t), \mathbf{w}\rangle + \frac{1}{2\eta_t} \lVert\mathbf{w} - \mathbf{w}_t\rVert_2^2,
    \end{aligned}
    \label{OMD-2norm}
\end{equation}
\end{shrinkeq}
where $\nabla  {f}_t(\mathbf{w}_t)$ is the subgradient of $ {f}_t$ at point $\mathbf{w}_t$ and $\eta_t$ is the stepsize at time $t$. In our problem, $\nabla  {f}_t(\mathbf{w}_t)$ can be calculated as:
\begin{shrinkeq}{-1ex}
\begin{equation}
    \begin{aligned}
    \nabla  {f}_t(\mathbf{w}_t) = 2\bar{\mathbf{z}}_t + 4\beta \mathbf{w}_t - \alpha\mathbf{S}^{\top}(\mathbf{S}\mathbf{w}_t)^{.(-1)}, 
    \end{aligned}
    \label{gradient of ft of wt}
\end{equation}
\end{shrinkeq}
and $.(-1)$ is an element-wise operator. 
By the optimal condition of \eqref{OMD-2norm}, we reach that
\begin{shrinkeq}{-1ex}
\begin{equation}
    \begin{aligned}
     \breve{\mathbf{w}}_{t} = \underset{\mathcal{W}_v}{\Pi}(\mathbf{w}_t - \eta_t\nabla  {f}_t(\mathbf{w}_t)),
    \end{aligned}
    \label{OMD-2norm-update}
\end{equation}
\end{shrinkeq}
where $\underset{\mathcal{W}_v}{\Pi}(\cdot)$ means mapping variables  into the set $\mathcal{W}_v$.

The core problem now becomes how to integrate evolutionary patterns \eqref{state transition function} into the above online framework. The way we leverage $\Phi$ is to perform one more update on the basis of \eqref{OMD-2norm-update}. Specifically, after obtaining $ \breve{\mathbf{w}}_{t}$, we use dynamic model $\Phi$ to update the graph of the next time period $\mathbf{w}_{t+1}$, i.e.,
\begin{shrinkeq}{-1ex}
\begin{equation}
    \begin{aligned}
    \mathbf{w}_{t+1} = \Phi( \breve{\mathbf{w}}_{t}).
    \end{aligned}
    \label{dynamic predict}
\end{equation}
\end{shrinkeq}
We can actually regard \eqref{dynamic predict} as a prediction step to output the graph of $t+1$ using prior $\Phi$. The extra prediction step brings more information beside gradient calculated from data and hence boosts learning performance. However, we should mention that an explicit prior $\Phi$ is not always available. 

In scenarios where no dynamic prior is available, inspired by \cite{simonetto2020time}, we employ a data-driven method to describe $\Phi$. The basic assumption is that graphs will evolve to the one satisfying the first order optimal condition of $ {f}_{t+1}$. Suppose  $\mathbf{w}_{t+1|t}$ is a prediction  of dynamic graph  at $t+1$ given information up to $t$ and we assume $\mathbf{w}_{t+1|t}$  satisfies \cite{simonetto2020time}:
\begin{shrinkeq}{-1ex}
\begin{equation}
    \begin{aligned}
        &\nabla_{\mathbf{w}}  {f}_{t+1}(\mathbf{w}_{t+1|t}) + \mathcal{N}_{\mathcal{W}_v} (\mathbf{w}_{t+1|t})  \ni \mathbf{0},
    \end{aligned}
    \label{1st order optimal condition}
\end{equation}
\end{shrinkeq}
where $\nabla_{\mathbf{w}}  {f}_{t+1}(\mathbf{w}_{t+1|t})$ is the partial derivative w.r.t. $\mathbf{w}$ and $\mathcal{N}_{\mathcal{W}_v} (\cdot)$ is the normal cone operator (the subdifferential of the indicator function) defined in \cite{simonetto2020time}. However, it is impossible to immediately solve \eqref{1st order optimal condition} since we have no information of $ {f}_{t+1}$ at $t$. An alternative way is to approximate \eqref{1st order optimal condition} by using the following method \cite{simonetto2017prediction},
\begin{shrinkeq}{-1ex}
    \begin{align}
        &\nabla_{\mathbf{w}}  {f}_{t+1}(\mathbf{w}_{t+1|t}) + \mathcal{N}_{\mathcal{W}_v} (\mathbf{w}_{t+1|t}) \notag\\[-0.2ex]
        \approx
        &\nabla_{\mathbf{w}}  {f}_t(\mathbf{\breve{\mathbf{w}}}_t) + \nabla_{\mathbf{w}\mathbf{w}}  {f}_t(\breve{\mathbf{w}}_t) (\mathbf{w}_{t+1|t} - \breve{\mathbf{w}}_{t})\notag\\[-0.2ex]
        &+ h \nabla_{t\mathbf{w}}  {f}_t(\breve{\mathbf{w}}_{t}) + \mathcal{N}_{\mathcal{W}_v} (\mathbf{w}_{t+1|t})\ni \mathbf{0},
    \label{1st order optimal condition prediction}    
    \end{align}
\end{shrinkeq}
where $\nabla_{\mathbf{w}\mathbf{w}}  {f}_t$ is the Hessian matrix of $ {f}_t$ w.r.t. $\mathbf{w}$, $\nabla_{t\mathbf{w}}  {f}_t$ is the  partial derivative of the gradient
of $ {f}_t$ w.r.t.  $t$ and $h$ is the  sample interval between two slots. In practice, $\nabla_{t\mathbf{w}}  {f}_t(\breve{\mathbf{w}}_t)$  is estimated by $         \left(\nabla_{\mathbf{w}}f_t(\breve{\mathbf{w}}_t) - \nabla_{\mathbf{w}}f_{t-1}(\breve{\mathbf{w}}_t)\right)/{h}$ \cite{simonetto2020time}. Observe  that \eqref{1st order optimal condition prediction} is equivalent to solving  the following problem
\begin{shrinkeq}{-1ex}
    \begin{align}
        &\mathbf{w}_{t+1|t}  \notag\\[-0.5ex]
        =& \underset{\mathbf{w}\in\mathcal{W}_v}{\mathrm{arg\;min}} \bigg{\{}\frac{1}{2} \mathbf{w}^{\top}\nabla_{\mathbf{w}\mathbf{w}} {f}_t(\breve{\mathbf{w}}_t)\mathbf{w} + \big{(}\nabla_{\mathbf{w}}  {f}_t(\breve{\mathbf{w}}_t)\notag\\[-1.5ex]
        &\hspace{1.5cm}+h \nabla_{t\mathbf{w}}  {f}_t(\breve{\mathbf{w}}_{t}) - \nabla_{\mathbf{w}\mathbf{w}}  {f}_t(\breve{\mathbf{w}}_t) \breve{\mathbf{w}}_t\big{)}^{\top}\mathbf{w}
        \bigg{\}}.
    \label{equivalent formulation of Taylor backward}
    \end{align}
\end{shrinkeq}
This equation can actually be regarded as a prediction of graphs in next time slot given $\breve{\mathbf{w}}_t$. Therefore, it is able to describe $\Phi$ when no explicit pattern is available. 

Instead of calculating the exact solution of \eqref{equivalent formulation of Taylor backward}, which may bring high computational cost, we try to find an approximate solution $\hat{\mathbf{w}}_{t+1|t}$   \cite{simonetto2017prediction}. Specifically, we set a dummy variable initialized as $\Tilde{\mathbf{w}}^0 = \breve{\mathbf{w}}_t$ and the following update steps are performed iteratively,
\begin{shrinkeq}{-1ex}
    \begin{align}
       \Tilde{\mathbf{w}}^{k+1} = \underset{\mathcal{W}_v}{\Pi} \big{(}&\Tilde{\mathbf{w}}^{k} - a (\nabla_{\mathbf{w}\mathbf{w}} {f}_t(\breve{\mathbf{w}}_t)(\Tilde{\mathbf{w}}^{k} - \breve{\mathbf{w}}_t) \notag\\[-1ex]
       &+ h\nabla_{t\mathbf{w}}{f}_t(\breve{\mathbf{w}}_t) + \nabla_{\mathbf{w}} {f}_t(\breve{\mathbf{w}}_t) )\big{)},
    \label{prediction step}
    \end{align}
\end{shrinkeq}
for $k = 0,...,K-1$, where $K$ is the pre-determined number of iterations and $a$ is a stepsize. After $K$ updates, we have
\begin{shrinkeq}{-1ex}
\begin{equation}
    \begin{aligned}
        \hat{\mathbf{w}}_{t+1|t} = \Tilde{\mathbf{w}}^{K}.
    \end{aligned}
    \label{estimate x_t}
\end{equation}
\end{shrinkeq}
The estimated $\hat{\mathbf{w}}_{t+1|t}$ is taken as the predicted $\mathbf{w}_{t+1}$.

The prediction step \eqref{dynamic predict} of our algorithm brings extra computational burden inevitably. For explicit dynamic priors, the extra computation is not heavy but brings better learning performance as discussed in experimental section. On the other hand, more computational burden may be incurred for the data-driven $\Phi$, especially for large $K$. However, it still provides a way to construct $\Phi$ without explicit dynamic priors.


\begin{algorithm}[t]
    \caption{Online graph learning algorithm with prior dynamic models (OGLP)}
    \label{alg:2}
    \begin{algorithmic}[1]
    \REQUIRE ~~\\ 
    $\alpha$, $\beta$, $\gamma$,   $\eta_t$, $\mathbf{z}_t$ for $t=1,...T$
    \ENSURE ~~\\ 
    The learned graph sequence $\mathbf{w}_t$
    \STATE Initialize $\mathbf{w}_1$, $\bar{\mathbf{z}}_0 = \mathbf{0}$\\
        \FOR{ $t = 1,..,T$}
            \STATE Receive data $\mathbf{x}_t$
            \STATE Update $\bar{\mathbf{z}}_t =\gamma \bar{\mathbf{z}}_{t-1} +(1-\gamma) \mathbf{z}_t$
            \STATE Calculate the gradient $\nabla  {f}_t(\mathbf{w}_t)$ with $\bar{\mathbf{z}}_t$
            \STATE Update $ \breve{\mathbf{w}}_{t} = \underset{\mathcal{W}_v}{\Pi} \left(\mathbf{w}_t - \eta_t \nabla  {f}_t(\mathbf{w}_t)\right) $
            \STATE Update $$
                \mathbf{w}_{t+1} = \Phi( \breve{\mathbf{w}}_{t})$$
        \ENDFOR
    \end{algorithmic}
\end{algorithm}

\section{Dynamic Regret Analysis}
\label{sec:regret analysis}
Dynamic regret is a commonly used metric to evaluate  performance of online algorithm in dynamic environments \cite{hall2015online}. In our problem, it is defined as 
\begin{shrinkeq}{-1ex}
\begin{equation}
    \begin{aligned}
    Reg_d(T)  \triangleq \sum_{t=1}^T  {f}_t(\mathbf{w}_t)  - \sum_{t=1}^T {f}_t(\mathbf{w}^*_t),
    \end{aligned}
    \label{def:dynamic regret}
\end{equation}
\end{shrinkeq}
where $\mathbf{w}_t^* =  \underset{\mathbf{w}\in\mathcal{W}_v}{\mathrm{argmin}}\, {f}_t(\mathbf{w})$ is the instantaneous optimal solution. Equation \eqref{def:dynamic regret} indicates that dynamic regret quantifies  the cumulative loss incurred by an online algorithm relative to the loss corresponding
to the optimal instantaneous solutions. An online algorithm is admissible only if it yields a sublinear regret since online algorithms with sublinear regret perform asymptotically as well as the batch algorithm on average \cite{zaman2020online}. In this section, the dynamic regret of our online graph learning algorithm is performed.

\subsection{Basic Assumptions}

We first introduce some technical assumptions that are essential to dynamic regret analysis.

\emph{a1.} All data received are bounded, i.e., there exists a constant $B_z>0$ such that $\lVert\mathbf{z}_t\rVert_2 \leq B_z $ for all $t$.

\emph{a2.} There  is  no  isolated node  in  the  graphs  we  update, i.e., there exist a constant $deg_{\mathrm{min}}>0$ such that $\mathbf{S}\mathbf{w}_t\geq deg_{\mathrm{min}} \mathbf{1} $ (entry-wise inequality) for all $t$.

\emph{a3.} The updated $\mathbf{w}_t$ is bounded, i.e., there exists a constant $w_{max}$ such that $\mathbf{w}_t \leq w_{max}\mathbf{1}$ (entry-wise inequality)  for all $t$.

\emph{a4.} Dynamic model $\Phi$ is a contractive mapping, i.e., there exists $r\in (0,1]$ such that 
\begin{shrinkeq}{-1ex}
\begin{equation}
  \lVert \Phi (\mathbf{w}_1) - \Phi (\mathbf{w}_2)\rVert_2 \leq r\lVert \mathbf{w}_1 - \mathbf{w}_2\rVert_2, 
\end{equation}
\end{shrinkeq}
for all $\mathbf{w}_{1}, \mathbf{w}_{2} \in \mathcal{W}_v$.

Without loss of generality, \emph{a1} holds true naturally in real-world applications. As stated in \cite{saboksayr2021online}, $\nabla f_t(\mathbf{w})$ is a Lipschitz-continuous function with constant $4\beta + 2\alpha (d-1)/(\min\left(\mathbf{S}\mathbf{w}\right))^2$. If the selected stepsize satisfies $$\eta_t \leq \left(2\beta + \alpha (d-1)/(\min\left(\mathbf{S}\mathbf{w}_t\right))^2\right)^{-1},$$ the $f_t(\mathbf{w}_{t+1})$ is bounded, indicating that $\mathbf{w}_{t+1}$ is bounded and $\mathbf{S}\mathbf{w}_{t+1}>0$. For a given feasible initial $\mathbf{w}_1$, we can hence obtain a sequence of  $\mathbf{w}_t$ such that  \emph{a2} and \emph{a3}  are justified. The last assumption is made to avoid that  poor  prediction  of dynamic models will be exacerbated by  repeated update \cite{hall2015online}.

\subsection{Dynamic Regret of OGLP}
We first build the bound of the gradient $\nabla  {f}_t(\mathbf{w}_t)$.

\begin{proposition}
\vspace{-1ex}
Under Assumption a1-a3, there exists a constant $ {L}>0$ such that $\lVert \nabla  {f}_t(\mathbf{w}_t)\rVert_2 \leq  {L}$ for all $\mathbf{w}_t\in\mathcal{W}_v$.
\label{proposition: gradient bound}
\vspace{-3ex}
\end{proposition}

\emph{Proof of Proposition \ref{proposition: gradient bound}}. 
Clearly,
\begin{shrinkeq}{-1ex}
\begin{equation}
    \begin{aligned}
        &\bar{\mathbf{z}}_t = \gamma \bar{\mathbf{z}}_{t-1} +(1-\gamma)\mathbf{z}_t\\
        =& (1-\gamma)[\gamma^{t-1} \mathbf{z}_1 + \gamma^{t-2} \mathbf{z}_2+...+\mathbf{z}_t]\\
        =&(1-\gamma)\sum_{\tau =1}^t\gamma^{t-\tau}\mathbf{z}_{\tau}.
    \end{aligned}
    \label{update z_t}
\end{equation} 
\end{shrinkeq}
We can thereby represent  $\nabla  {f}_t(\mathbf{w}_t)$ as
\begin{shrinkeq}{-1ex}
\begin{equation}
    \begin{aligned}
    \nabla  {f}_t(\mathbf{w}_t)& = 2(1-\gamma)\sum_{\tau =1}^t\gamma^{t-\tau}\mathbf{z}_{\tau} +4\beta\mathbf{w}_t -\alpha \mathbf{S}^{\top}(\mathbf{S}\mathbf{w}_t)^{.(-1)}
    \end{aligned}.
    \label{rewrite f_tilde_t}
\end{equation}
\end{shrinkeq}
Then for $t=1,...,T$ and $\gamma>0$,
\begin{shrinkeq}{-1ex}
    \begin{align}
     &\lVert\nabla  {f}_t(\mathbf{w}_t)\rVert_2 \notag\\[-0.5ex]
   \leq & 2(1-\gamma)\left\lVert \sum_{\tau =1}^t\gamma^{t-\tau}\mathbf{z}_{\tau} \right\rVert_2 +4\beta\left\lVert \mathbf{w}_t\right\rVert_2 + \alpha\left\lVert \mathbf{S}(\mathbf{S}\mathbf{w}_t)^{.(-1)}\right\rVert_2 \notag\\[-0.5ex]
     \leq & 2(1-\gamma)\left( \sum_{\tau =1}^t\gamma^{t-\tau} B_z\right ) +4\beta\left\lVert \mathbf{w}_t\right\rVert_2 + \alpha\left\lVert \mathbf{S}(\mathbf{S}\mathbf{w}_t)^{.(-1)}\right\rVert_2 \notag\\[-0.5ex]
    \leq &2B_z(1 - \gamma^t) + 4\beta\left\lVert \mathbf{w}_t\right\rVert_2 + \alpha\left\lVert\mathbf{S} \right\rVert_2\left\lVert (\mathbf{S}\mathbf{w}_t)^{.(-1)}\right\rVert_2 \notag\\[-0.1ex]
    \leq &2B_z(1 - \gamma^t) +2\sqrt{2}\beta\sqrt{d(d-1)}w_{\mathrm{max}} \notag\\[-0.5ex]
    & \; + \alpha \sqrt{2(d-1)}\frac{\sqrt{d}}{deg_{\mathrm{min}}} \notag\\[-1ex]
    \leq &2B_z+2\sqrt{2}\beta\sqrt{d(d-1)}w_{\mathrm{max}} + \alpha \sqrt{2(d-1)}\frac{\sqrt{d}}{deg_{\mathrm{min}}} \notag\\[-1ex]
     \triangleq&  {L},
    \label{norm of rewrite f_tilde_t}
    \end{align}
\end{shrinkeq}
The second inequality holds because of assumption \emph{a1}. The fourth inequality holds because for all $\mathbf{w}_t\in \mathcal{W}_v$,
$\lVert \mathbf{w}_t\rVert_2 \leq \sqrt{\frac{d(d-1)}{2}}w_{\max}$ under assumption \emph{a3} and
$\lVert\mathbf{S}\rVert_2 = \sqrt{2(d-1)}$ (cf. \cite{SABOKSAYR2021108101}). In addition, $\lVert(\mathbf{S\mathbf{w}}_t)^{.(-1)}\rVert_2\leq \frac{1}{{deg}_{\mathrm{min}}}\lVert\mathbf{1}\rVert_2 = \frac{\sqrt{d}}{deg_{\mathrm{min}}}$ due to assumption \emph{a2}.

\begin{figure*}[t] 
    \centering
	  \subfloat[]{
       \includegraphics[width=0.32\textwidth]{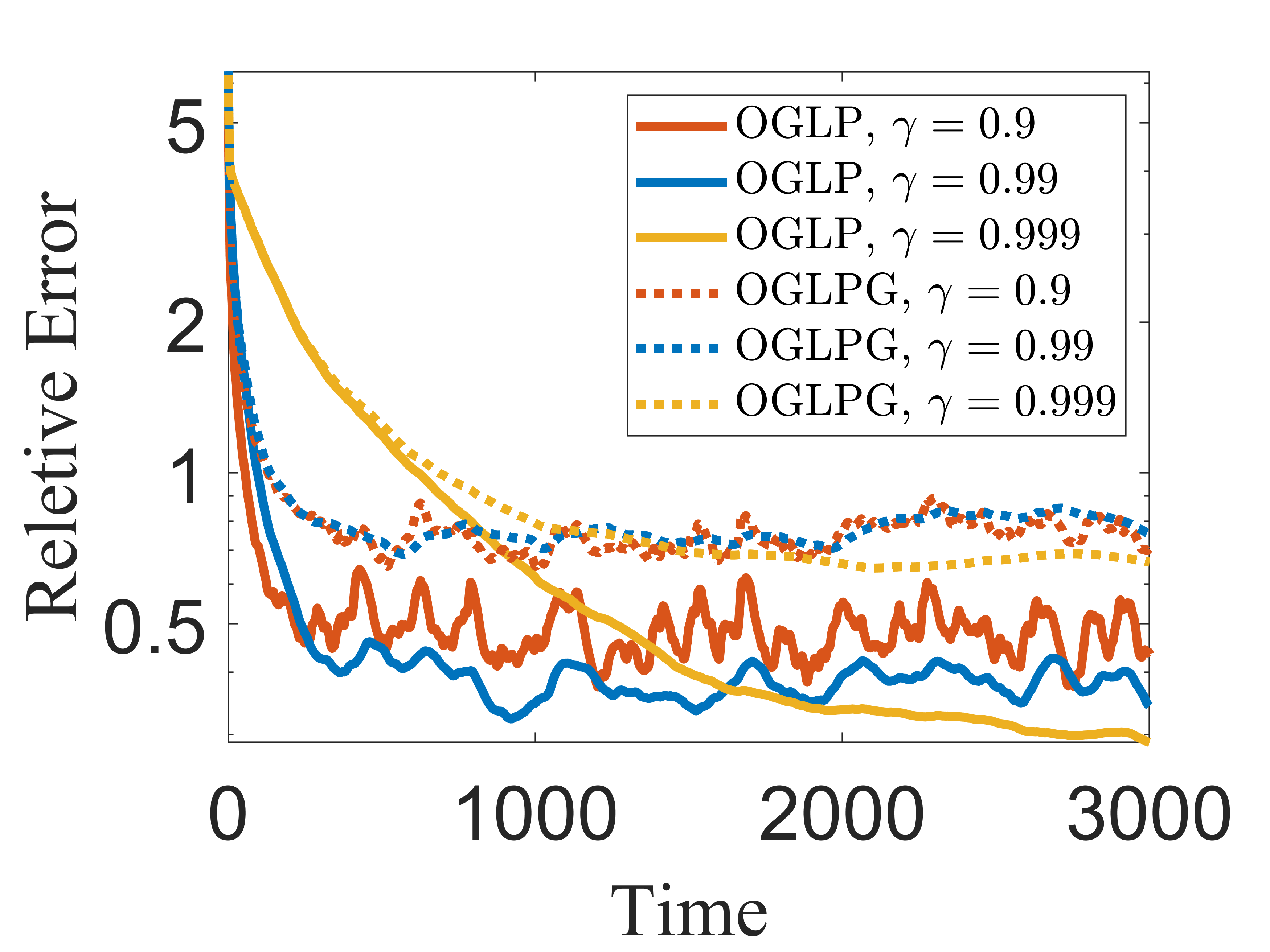}}
	  \hspace{15mm}
	  \subfloat[]{
        \includegraphics[width=0.32\textwidth]{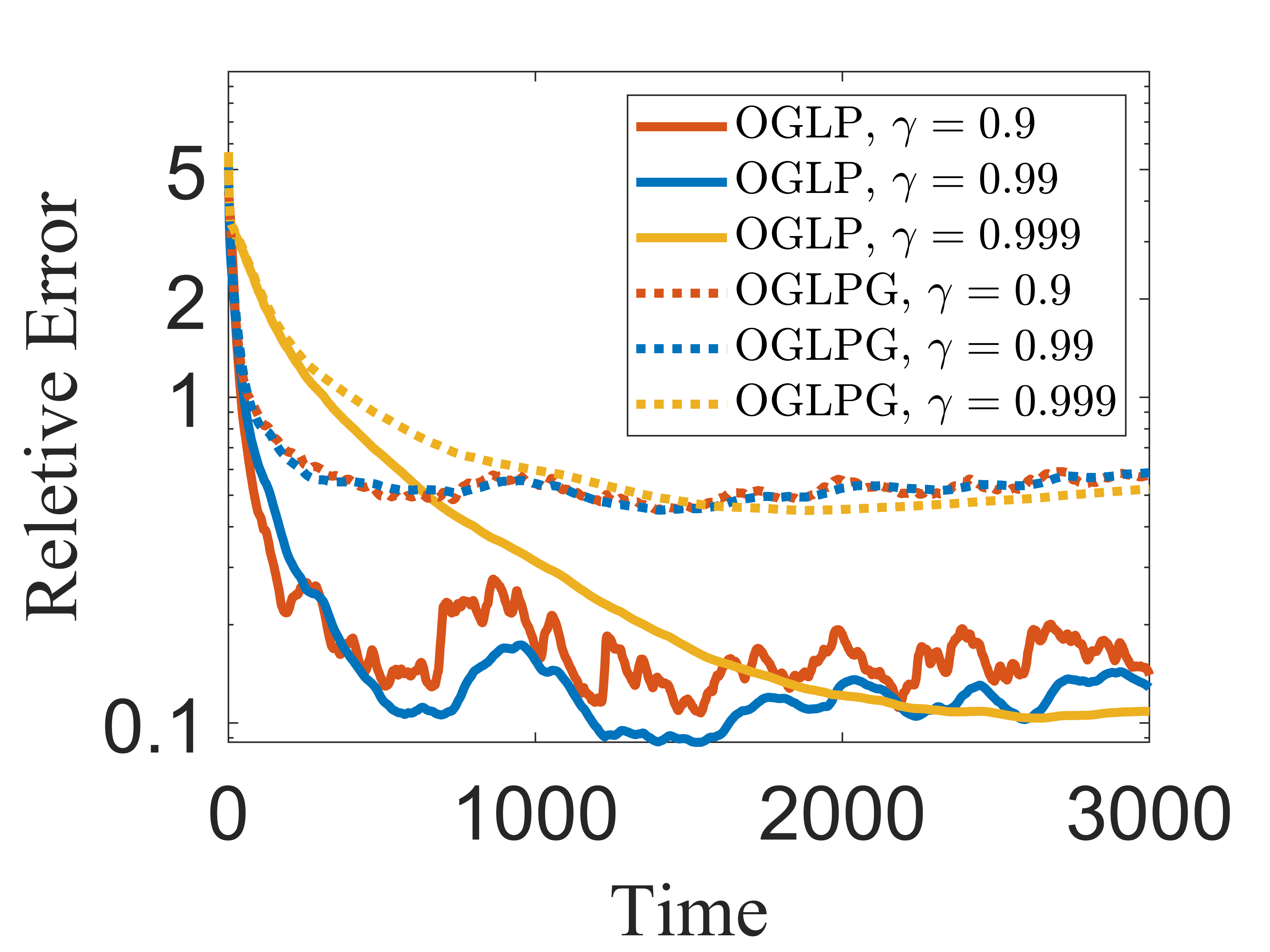}}
       \hspace{15mm}
      \subfloat[]{
        \includegraphics[width=0.32\textwidth]{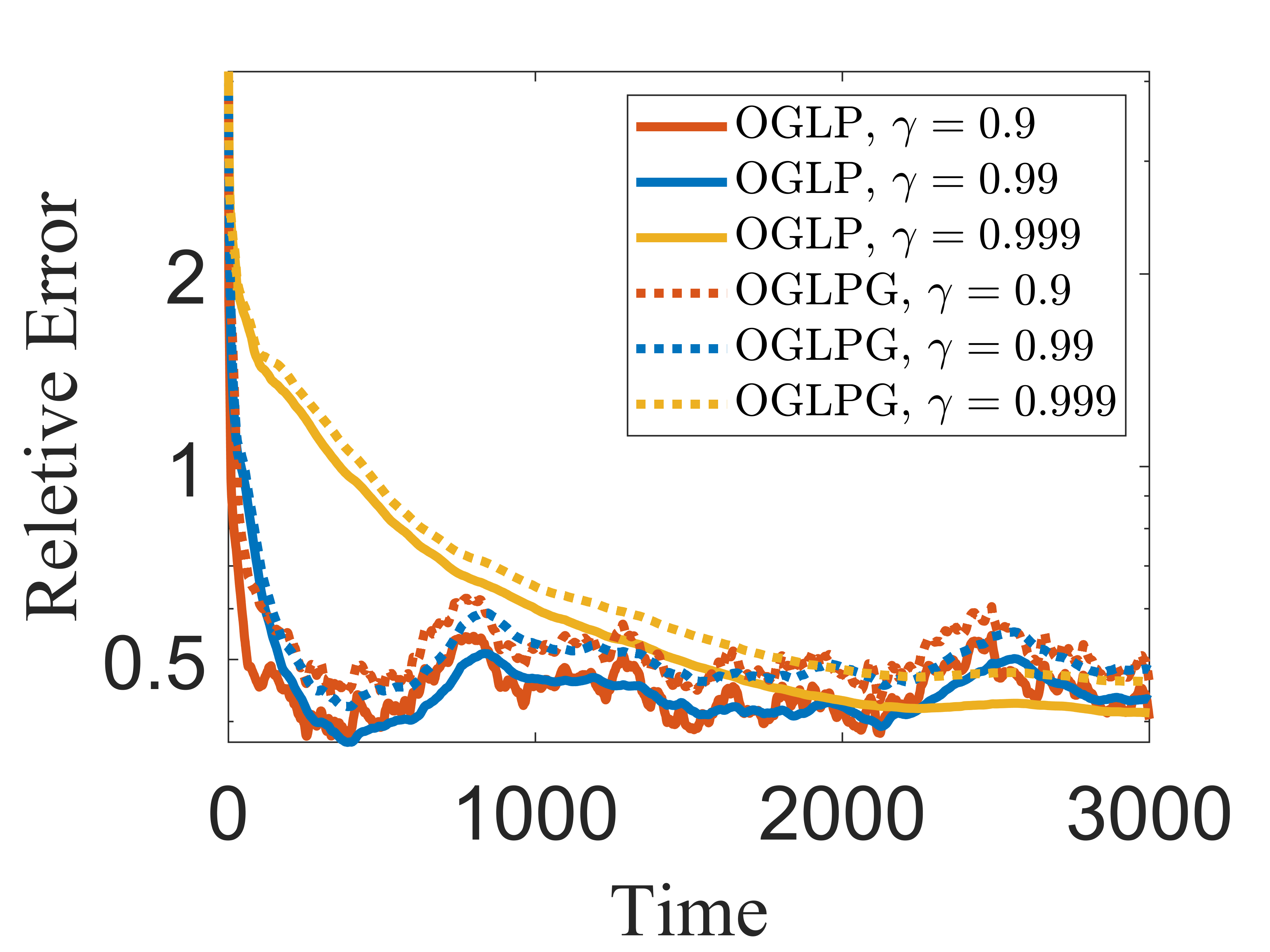}}
        \hspace{15mm}
       \subfloat[]{
        \includegraphics[width=0.32\textwidth]{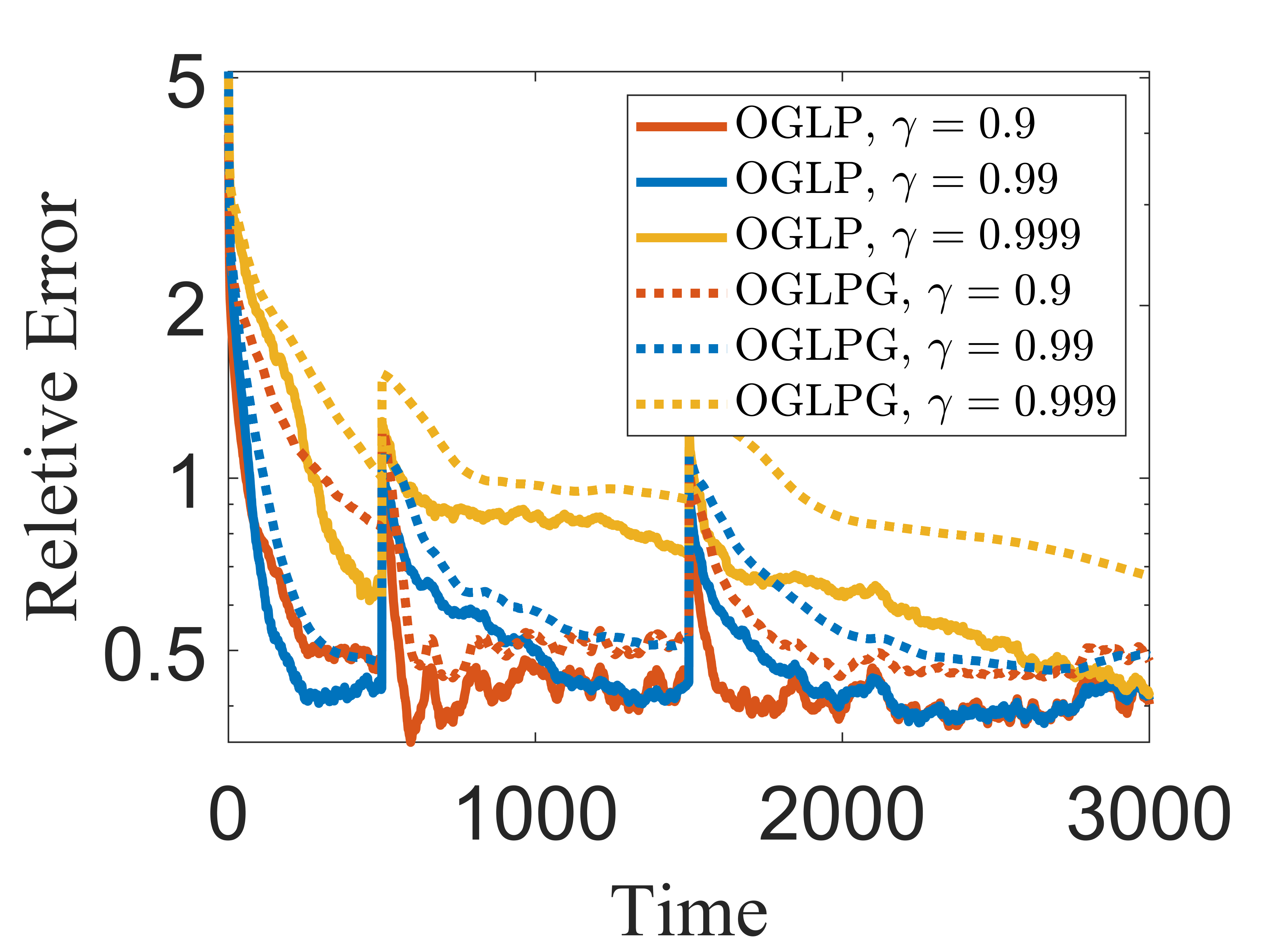}}
    	\caption{ Relative error of different dynamic models (a) AR model (b) Transition model (c) Social network  (d) Switching model}
    	\label{Fig.error}
    	\vspace{-3ex}
\end{figure*}

We next provide the  dynamic regret of OGLP algorithm. 
\begin{theorem}
Under assumption a1- a4,  if the selected stepsize $\eta_t$  is a constant satisfying $\eta_t = \eta \leq \left(2\beta + \alpha(d-1)/deg_{\mathrm{min}}^2\right)^{-1}$, the dynamic regret $Reg_d$ of OGLP algorithm satisfies
\begin{shrinkeq}{-1ex}
\begin{equation}
    \begin{aligned}
        &Reg_d(T)  = \sum_{t=1}^T{f}_t(\mathbf{w}_t)  -  \sum_{t=1}^T {f}_t(\mathbf{w}^*_t) \\
        \leq &\frac{d(d-1)w_{\mathrm{max}}^2}{4\eta} +\frac{\sqrt{2d(d-1)}w_{\mathrm{max}}}{2\eta} C_V^d+ \frac{\eta T}{2} {L}^2,
    \end{aligned}
    \label{regret_u}
\end{equation}
\end{shrinkeq}
where $C_V^d \triangleq \sum_{t=2}^T \lVert \mathbf{w}^*_{t} - \Phi(\mathbf{w}^*_{t-1})\rVert_2$.
\label{theom: regret_u}
\end{theorem}

\emph{Proof of Theorem \ref{theom: regret_u}}. The sketch of the proof is derived from \cite{zinkevich2003online} and can also be found in \cite{zhang2018adaptive}. However, we made  some minor modifications to the proof for our problem. The first one is the bound of $\lVert\nabla  {f}_t(\mathbf{w}_t) \rVert_2$ used in the proof procedure is the $ {L}$  in Proposition \ref{proposition: gradient bound}. The other modification is that the upper bound of $\lVert \mathbf{w}_t \rVert_2$ for $\mathbf{w}_t\in \mathcal{W}_v$ is $\sqrt{\frac{d(d-1)}{2}}w_{\max}$.

From the definition of $C_V^d$, if $\Phi$ is in line with the real evolutionary patterns of dynamic graphs, we can conclude that $C^d_V \leq C_V \triangleq \sum_{t = 2}^T \lVert \mathbf{w}^*_t - \mathbf{w}^*_{t-1}\rVert_2$, where $C_V$ measures the variations of dynamic environments. Therefore, the prediction step using dynamic models $\Phi$ theoretically reduces the impact of dynamic environments, bringing a tighter regret bound compared with those without  priors. Furthermore,
if the selected  $\eta$ satisfies $\eta = \mathcal{O}(1/\sqrt{T})$,   our algorithm can actually achieve $\mathcal{O}(\sqrt{T}(1 + C_V^d))$  regret.  Finally, note that the constant stepsize is required to be smaller than $\left(2\beta + \alpha (d-1)/deg_{\mathrm{min}}^2\right)^{-1}$. However, $deg_{\mathrm{min}}$ is a constant we cannot obtain in advance. An alternative is to use an adaptive stepsize introduced in \cite{saboksayr2021online}, which is also adopted in our experiment.

\section{Numerical Experiments}
\label{sec:numerical experiments}
\subsection{Synthetic data}
We test our method on four dynamic models listed below.

\emph{1).} Autoregressive (AR) model \cite{hall2015online}. A first-order AR model is used
\begin{shrinkeq}{-1ex}
\begin{equation}
    \begin{aligned}
        \mathbf{w}_{t+1} = \mathbf{A}\mathbf{w}_t,
    \end{aligned}
    \label{pattern-AR}
\end{equation}
\end{shrinkeq}
where $\mathbf{A}\in \mathbb{R}^{p\times p}$ is an autoregressive transition matrix.

\emph{2).} Transition model. In this model, graphs will evolve to a target topology  known in advance, i.e, 
\begin{shrinkeq}{-1.2ex}
\begin{equation}
    \begin{aligned}
       \mathbf{w}_{t+1} = a\mathbf{w}_{t} + (1-a) \mathbf{w}_{\mathrm{target}},
    \end{aligned}
    \label{pattern-transition}
\end{equation}
\end{shrinkeq}
where $0<a<1$ is a predefined controlling parameter and $\mathbf{w}_{\mathrm{target}}$ is the target graph. 

\emph{3).} Social network model. This model describes the evolution of social networks.  More details are referred in \cite{snijders2001statistical}, and we will not introduce it in depth  due to space limitation.

\emph{4).} Switching model. In this model, graphs switch to a new topology at some specific moments and remain static at other moments.

We first generate a graph following the way of \cite{dong2016learning} and take it as the initial graph at $t=1$. Graphs of the remaining time slots are  obtained by $\mathbf{w}_{t+1} = \Phi(\mathbf{w}_t)$ with $\Phi$ introduced above. All parameters of these dynamic models are carefully selected to prevent the situation where $\mathbf{w}_t\notin \mathcal{W}_v$. For switching model, we transfer graphs to some random topologies at $t = 500$ and $t=1500$. Smooth graph signal $\mathbf{x}_t$ is generated from graph at $t$ by the same way introduced in \cite{dong2016learning}. We set $T = 3000$ and the obtained signals are taken as inputs sequentially. The adopted evaluation metric   is relative error representing  the accuracy of edge weight, i.e., $\lVert {\mathbf{W}}_t - \mathbf{W}^*_t\rVert_{\text{F}}/{\lVert \mathbf{W}^*_t\rVert_{\text{F}}}$, where $ {\mathbf{W}}_t$ is the adjacency matrix transformed  from $\mathbf{w}_t$ and $\mathbf{W}^*_t$ is the groundtruth of $t$. In our experiments, we fix $\alpha = 2$ and search the best $\beta$ as \cite{kalofolias2016learn} does. Since we have no knowledge of $deg_{\mathrm{min}}$ in advance, we hence adopt the adaptive stepsize $\eta_t = \left(2\beta + \alpha (d-1)/\mathrm{min}(\mathbf{S}\mathbf{w}_t)^2 \right)^{-1}$ as \cite{saboksayr2021online} does. Furthermore,  we  take \cite{saboksayr2021online}  as  baseline  and  name  it as OGLPG  since  it  is  the  only  method  we  can  find   in  the literature  that  learns graphs  under  smoothness  prior  in  an  online  fashion. The parameters of OGLPG is the same as ours.

Figure \ref{Fig.error} shows the performance of the above  models. For the same $\gamma$, our method with dynamic priors outperforms OGLPG since lower relative errors are obtained as rounds increase. Furthermore, our method exhibits stronger tracking ability. For switching model, fewer iterations are required to restore previously low relative error. Note that there exists no explicit form of $\Phi$ in switching model. Hence we adopt the data-driven method introduced in section \ref{sec:Proposed Framework} to construct $\Phi$. Another interesting point is that large $\gamma$ results in a ``smoother" learning curve by sacrificing tracking performance while small $\gamma$ is more susceptible to new data. More oscillations occur in the curves of $\gamma = 0.9$. Therefore, it is a trade-off between tracking ability and stability to choose a suitable $\gamma$.

\begin{figure}[t]
    \centering
       \includegraphics[width=1\linewidth]{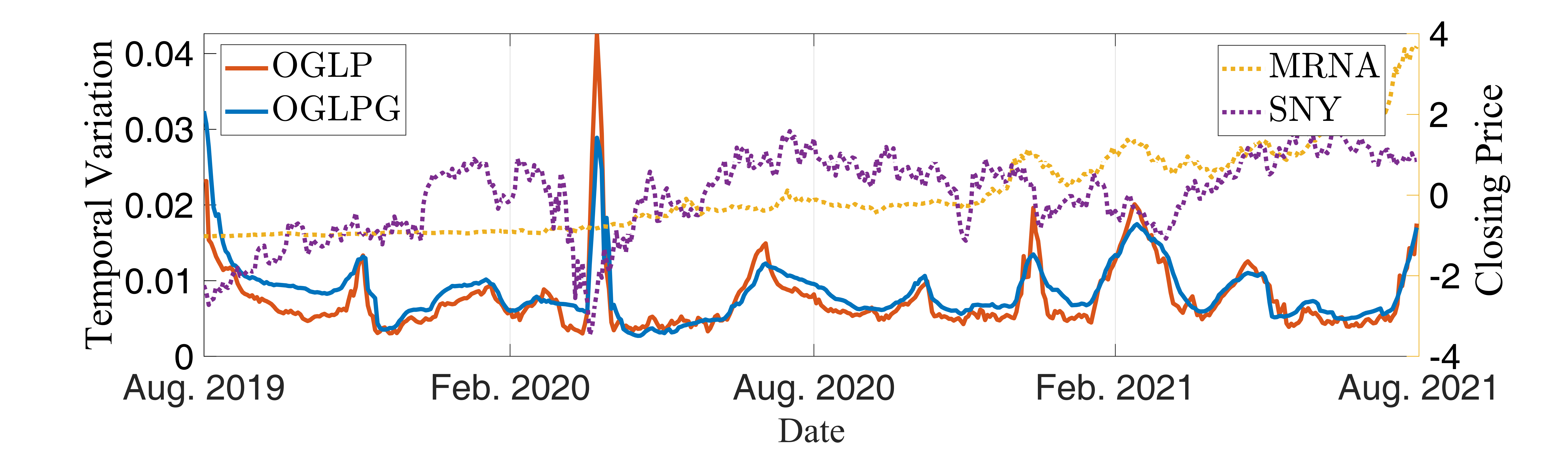}
    	\caption{Temporal variation of stock market graph}
	    \label{fig-stock}
	    \vspace{-1em}
\end{figure}

\subsection{Real data}
Our method is also applied to stock closing price data\footnote{https://finance.yahoo.com/} to learn time-varying graphs of relationships  among different companies. We collect stock prices data of 10 pharmaceutical companies, including Amgen (AMGN), Astrazeneca (AZN), GlaxoSmithKline (GSK), Johnson \& Johnson (JNJ), Moderna (MRNA), Novavax (NVAX), Pfizer (PFE), Perrigo (PRGO), Sanofi (SNY) and Zoetis (ZTS). We focus on data from August 1st 2019 to July 30th 2021, during  which COVID-19 pandemic outbroke and caused market
instabilities of pharmaceutical companies. All data are standardized first and we aim to detect  changes of the learned graphs in this unstable market. We set $T = 504$ (the number of working days from August 1st 2019 to July 30th 2021) and $d = 10$. Moreover,  $\alpha, \beta$ and $\gamma$ are set to be $2$, $1.2$, $0.99$ respectively for both OGLP and OGLPG. The data-driven method is used to predict graph evolution. We employ a metric named temporal variation $\lVert \mathbf{w}_{t} - \mathbf{w}_{t-1} \rVert_2/\lVert \mathbf{w}_{t-1} \rVert_2$ to  measure structural changes between two consecutive graphs. In Fig. \ref{fig-stock},  historical closing stock prices of two companies, i.e., MRNA and SNY, are first depicted to show the instability of the market. Compared with MRNA, the stock prices of SNY witnessed marked volatility in March, implying changes of the corresponding graph of relationship. As shown in Fig. \ref{fig-stock}, a significant temporal variation is detected in March 2020, which is consistent with changes of stock prices in March. However, OGLP tends to be more sensitive to changes, meaning better tracking ability in dynamic environments. When the market stabilizes, OGLP is able to return to the steady state (small temporal variations) more quickly.

\section{Conclusion}
\label{sec:conclusion}
In this paper, a framework of  learning graphs in an online fashion with dynamic priors is proposed. Where no priors are available, we use a data-driven method to predict evolution of dynamic graphs. Theoretical analysis illustrates that our online algorithm is able to reach sublinear dynamic regret bound. Experimental results show the superiority of our method.


\end{document}